\documentclass[11pt, a4paper]{article}

\usepackage[utf8]{inputenc}
\usepackage{fontenc}
\usepackage{amsmath}
\usepackage{amsfonts}
\usepackage{amssymb}
\usepackage{graphicx}
\usepackage{booktabs}
\usepackage{multicol}
\usepackage[margin=1in]{geometry}
\usepackage{hyperref}
\usepackage{url}
\usepackage{authblk}

\hypersetup{
    colorlinks=true,
    linkcolor=blue,
    filecolor=magenta,      
    urlcolor=cyan,
    pdftitle={Sparse-Query-Attention},
    pdfpagemode=FullScreen,
}

\title{
    \textbf{Sparse Query Attention (SQA): A Computationally Efficient Attention Mechanism with Query Heads Reduction}
}
\author{Adam Filipek (adamfilipek@rxai.dev)}
\affil{Reactive AI (https://rxai.dev)}
\date{September 2025}

\begin{document}

\maketitle

\begin{abstract}
The Transformer architecture, underpinned by the Multi-Head Attention (MHA) mechanism, has become the de facto standard for state-of-the-art models in artificial intelligence. However, the quadratic computational complexity of MHA with respect to sequence length presents a significant barrier to scaling, particularly for applications involving long contexts. Prevailing solutions, such as Multi-Query Attention (MQA) and Grouped-Query Attention (GQA), have effectively addressed the memory bandwidth bottleneck that dominates autoregressive inference latency by sharing Key and Value projections. While highly successful, these methods do not reduce the fundamental number of floating-point operations (FLOPs) required for the attention score computation, which remains a critical bottleneck for training and full-sequence processing. This paper introduces Sparse Query Attention (SQA), a novel attention architecture that pursues an alternative and complementary optimization path. Instead of reducing Key/Value heads, SQA reduces the number of \textit{Query} heads. This architectural modification directly decreases the computational complexity of the attention mechanism by a factor proportional to the reduction in query heads, thereby lowering the overall FLOPs. This work presents the theoretical foundation of SQA, its mathematical formulation, and a family of architectural variants. Empirical benchmarks on long sequences (32k-200k tokens) demonstrate that SQA can achieve significant throughput improvements of up to 3x in computation-bound scenarios such as model pre-training, fine-tuning, and encoder-based tasks, with only a minimal impact on model quality in preliminary small-scale experiments. SQA was discovered serendipitously during the development of the upcoming Reactive Transformer architecture, a context in which its computational advantages are maximized, suggesting its potential as a powerful tool for building more efficient and scalable models.
\end{abstract}

\section{Introduction}

\subsection{The Computational Burden of Self-Attention}
The introduction of the Transformer architecture by Vaswani et al. (2017) marked a paradigm shift in sequence modeling and has since become the foundational building block for most modern large language models (LLMs) and other advanced AI systems. The source of the Transformer's remarkable power lies in its self-attention mechanism, which allows the model to dynamically weigh the importance of all tokens in a sequence relative to each other, capturing complex, long-range dependencies without the sequential constraints of recurrent networks.

This representational power, however, comes at a steep price. The core of the self-attention mechanism involves computing a dot product between a matrix of queries ($Q$) and the transpose of a matrix of keys ($K$), an operation whose computational and memory requirements scale quadratically with the sequence length, $N$. The computational complexity of the standard Multi-Head Attention (MHA) is formally expressed as $O(N^2 \cdot d_{\text{model}} + N \cdot d_{\text{model}}^2)$, where $d_{\text{model}}$ is the model's hidden dimension. For the long sequences that are increasingly common in modern applications, the $N^2$ term becomes the dominant factor, making self-attention a formidable computational bottleneck. This quadratic scaling limits the context length that models can feasibly process, hindering progress in areas such as long-document understanding, extended dialogue, and high-resolution multimodal processing. Consequently, the development of more efficient attention mechanisms has become one of the most critical areas of research in deep learning.

\subsection{A Tale of Two Bottlenecks: Computation vs. Memory Bandwidth}
To develop effective optimizations, it is essential to recognize that the performance of Transformer models is constrained by two distinct, though related, bottlenecks. The failure to differentiate between these two challenges can lead to solutions that are highly effective in one scenario but suboptimal in another.

The first is the \textbf{computational bottleneck}, which refers to the sheer volume of floating-point operations (FLOPs) required to execute the attention algorithm. This is primarily dictated by the matrix multiplication $QK^T$, which involves $O(N^2 \cdot d_{\text{model}})$ operations. This bottleneck is most prominent in scenarios where computations can be heavily parallelized and the full sequence is processed at once. Such scenarios include:
\begin{itemize}
    \item \textbf{Model Pre-training and Fine-tuning:} During training, forward and backward passes are performed on large batches of long sequences, where the primary limitation is the raw throughput of the compute hardware (e.g., Tensor Cores on GPUs).
    \item \textbf{Encoder Architectures:} Models like BERT (Devlin, J., et al., 2019) or the encoder component of sequence-to-sequence models process the entire input sequence in a single, parallel step.
    \item \textbf{Prompt Processing in Decoders:} When a decoder-only LLM is given a long prompt, the initial processing of that prompt to generate the first token is a parallel operation on the entire prompt sequence, which is compute-bound.
\end{itemize}

The second is the \textbf{memory bandwidth bottleneck}. This issue is most acute during autoregressive decoding, the token-by-token generation process used by LLMs for inference. At each generation step, the model must compute attention between the query for the new token and the keys and values for \textit{all} previous tokens in the sequence. These past keys and values are stored in a Key-Value (KV) cache in high-bandwidth memory (HBM). The bottleneck arises from the need to load this entire, ever-growing KV cache from HBM into the much faster but smaller on-chip SRAM of the GPU for every single token that is generated. For long sequences, the size of the KV cache can reach several gigabytes, and the time spent on this data transfer can far exceed the time spent on actual computation, making inference latency memory-bound.

The research community's focus has disproportionately gravitated towards solving the memory bandwidth bottleneck, driven by the pressing need for low-latency inference in commercial LLM applications. This has led to groundbreaking innovations but has also created an environment where optimizations for the equally important computational bottleneck of training and encoding have been comparatively underexplored. This reveals a systemic bias toward a specific application profile—the interactive, decoder-only LLM—potentially leaving significant performance gains on the table for other critical use cases.

\subsection{Existing Paradigms: Optimizing for the Memory Bottleneck}
The dominant approaches to creating more efficient attention mechanisms, Multi-Query Attention (MQA) (Shazeer, 2019) and Grouped-Query Attention (GQA) (Ainslie et al., 2023), are masterful solutions designed explicitly to alleviate the memory bandwidth bottleneck.

MQA takes a radical approach by having all query heads share a single, common projection for keys and values, dramatically reducing the KV cache size. GQA provides a more nuanced interpolation between the standard MHA and the aggressive MQA. It divides the query heads into several groups and assigns a shared key/value projection to each group. This allows model architects to strike a balance, achieving most of the speed benefits of MQA while mitigating the potential quality degradation that can arise from having only a single key/value representation.

More recently, Multi-head Latent Attention (MLA), introduced in models like DeepSeek-V2 (DeepSeek-AI, 2024), represents a further evolution in this direction. MLA compresses the Key and Value tensors into a low-rank latent representation before they are cached. This technique achieves an even greater reduction in the KV cache size, pushing the boundaries of memory efficiency for autoregressive inference. The progression from MHA to MQA, GQA, and MLA illustrates a clear research trajectory focused on minimizing data movement from HBM. However, it is crucial to recognize their underlying mechanism: they optimize performance by reducing the \textit{size} of the data being transferred, not the \textit{amount of computation} performed. In all these methods, the number of query heads remains unchanged. As a result, the size of the query matrix $Q$ and the dimensions of the resulting attention score matrix remain the same as in MHA. Consequently, the number of FLOPs required for the $QK^T$ operation is not reduced. While they are indispensable for fast inference, they do not accelerate the compute-bound tasks of training or full-sequence encoding.

\subsection{The Broader Landscape of Efficiency}
Beyond memory-centric optimizations, two other major research directions have emerged to tackle the Transformer's scaling challenges: approximating full attention and developing entirely new architectural paradigms.

\subsubsection{Approximating Full Attention: Sliding Window Mechanisms}
A widely adopted technique to move from quadratic to linear complexity is Sliding Window Attention (SWA) (Beltagy et al., 2020). Instead of attending to all tokens in the sequence, each token only attends to a fixed-size local window of neighboring tokens. This reduces the computational complexity from $O(N^2)$ to $O(N \cdot k)$, where k is the window size. While highly effective, SWA's primary limitation is its inability to capture dependencies between tokens that are farther apart than the window size. To mitigate this, architectures like Longformer (Beltagy et al., 2020) combine SWA with a few designated "global" tokens that can attend to the entire sequence. SWA is a complementary mechanism, often used in conjunction with other attention variants like GQA in models such as Gemma (Google, 2024) and Mistral, to manage long contexts efficiently.

\subsubsection{Architectural Alternatives to the Transformer}
A more radical approach involves replacing the attention mechanism entirely with sub-quadratic alternatives. State Space Models (SSMs), exemplified by Mamba (Gu \& Dao, 2023), have emerged as a powerful alternative. Inspired by classical state space models from control theory, SSMs are designed to operate with linear complexity in sequence length, making them highly efficient for very long sequences. Similarly, Retentive Networks (RetNet) (Sun et al., 2023) derive a connection between recurrence and attention to achieve a parallelizable training process with linear-time inference. Other frameworks, such as Hierarchical Memory Transformers (HMT) (He et al., 2025), augment existing models with external memory and recurrent mechanisms to process sequences in chunks. These architectures represent a fundamental departure from the Transformer paradigm, offering significant performance benefits at the cost of moving away from the well-established attention-based ecosystem.

However, these alternatives remain niche due to notable challenges in scalability and training stability. SSMs, while theoretically linear, suffer from an "illusion of state" - their expressive power is limited similarly to Transformers, struggling with true long-distance dependencies and exhibiting RNN-like gradient issues on massive datasets (Deletang et al., 2024). RetNet's hybrid nature introduces locality biases that degrade performance in translation or reasoning tasks. HMT addresses long contexts via hierarchy but incurs compression overhead and hardware scalability bottlenecks, limiting adoption in production-scale LLMs (He et al., 2025). In contrast, optimizations like SQA evolve the attention layer itself, preserving Transformer's parallelism, mature tooling like, FlashAttention (Dao, T., et al., 2022), and representational strengths for tasks like encoder-based processing, while delivering constant-factor FLOPs reductions.

\subsection{Our Contribution: Sparse Query Attention (SQA)}
This paper introduces Sparse Query Attention (SQA), a novel attention mechanism that directly addresses the computational bottleneck by reducing the number of FLOPs. The core idea is simple yet counter-intuitive in the context of existing work: instead of a further reduction of the number of key and value heads, SQA reduces the number of \textit{query} heads.

This architectural change has a direct and profound impact on the computational graph. The complexity of the attention score calculation is proportional to the number of query heads. By reducing the number of query heads from $H$ (the total number of heads in a comparable MHA model) to $H_q$ (where $H_q < H$), SQA reduces the computational cost of the attention layer by a factor of $H / H_q$. This is not a memory optimization; it is a fundamental reduction in the amount of arithmetic required.

This work makes the following contributions to the field of efficient Transformer architectures:
\begin{itemize}
    \item It introduces Sparse Query Attention (SQA), a new attention mechanism that reduces computational complexity by a factor of $H / H_q$, where $H$ is the total number of heads in a baseline MHA model and $H_q$ is the number of query heads in the SQA model.
    \item It provides a rigorous mathematical formulation of SQA and a formal analysis of its computational and memory complexity profiles, contrasting them with MHA, MQA, and GQA.
    \item It presents a family of SQA variants, including Symmetric SQA (sSQA) and Extreme SQA (xSQA), which allow for exploration of the trade-off space between computational efficiency and model capacity.
    \item It empirically demonstrates through performance benchmarks that SQA achieves significant throughput improvements of up to 3x on long sequences (32k-200k tokens) in compute-bound scenarios, such as training and encoding, where MQA and GQA offer no speed advantage.
    \item It shows through preliminary, small-scale experiments that these substantial performance gains are achievable with only a minor impact on model quality, motivating the need for further large-scale research.
\end{itemize}

\section{Background: The Evolution of Efficient Attention}

\subsection{The Foundation: Multi-Head Attention (MHA)}
The original Multi-Head Attention mechanism, introduced as the cornerstone of the Transformer model, was designed to maximize representational power. The key idea is to allow the model to jointly attend to information from different representation subspaces at different positions. Instead of performing a single attention function, MHA runs multiple scaled dot-product attention operations, or "heads," in parallel and concatenates their results.

The core operation for each head is the scaled dot-product attention, defined as:
\begin{equation}
\text{Attention}(Q, K, V) = \text{softmax}\left(\frac{QK^T}{\sqrt{d_k}}\right)V
\label{eq:attention}
\end{equation}
Here, $Q$, $K$, and $V$ are the Query, Key, and Value matrices, respectively, and $d_k$ is the dimension of the keys. The scaling factor $\sqrt{d_k}$ is used to prevent the dot products from growing too large, which could push the softmax function into regions with extremely small gradients.

In MHA, the input representation is first linearly projected into queries, keys, and values for each of the $h$ heads. The output of each head is then concatenated and passed through a final linear projection. The full operation is defined as:
\begin{equation}
\text{MultiHead}(Q, K, V) = \text{Concat}(\text{head}_1, \dots, \text{head}_h)W^O
\end{equation}
where each head is computed as:
\begin{equation}
\text{head}_i = \text{Attention}(QW_i^Q, KW_i^K, VW_i^V)
\end{equation}
The matrices $W_i^Q$, $W_i^K$, $W_i^V$, and $W^O$ are learnable parameter matrices. This multi-headed structure allows each head to specialize and capture different types of relationships within the data, such as syntactic dependencies or long-distance semantic links.

The computational complexity of MHA is dominated by two main components: the matrix multiplication for the attention scores ($QK^T$) and the matrix multiplication for the output projection. For a sequence of length $N$ and a model dimension of $d_{\text{model}}$, with $h$ heads each of dimension $d_k = d_{\text{model}}/h$, the complexity of the score calculation is $O(h \cdot N^2 \cdot d_k) = O(N^2 \cdot d_{\text{model}})$. The complexity of the final projection is $O(N \cdot d_{\text{model}}^2)$. Thus, the total complexity is $O(N^2 \cdot d_{\text{model}} + N \cdot d_{\text{model}}^2)$. For long sequences where $N > d_{\text{model}}$, this is effectively $O(N^2 \cdot d_{\text{model}})$, establishing the quadratic dependency that motivates the search for more efficient alternatives.

\begin{figure}
    \centering
    \includegraphics[width=1\linewidth]{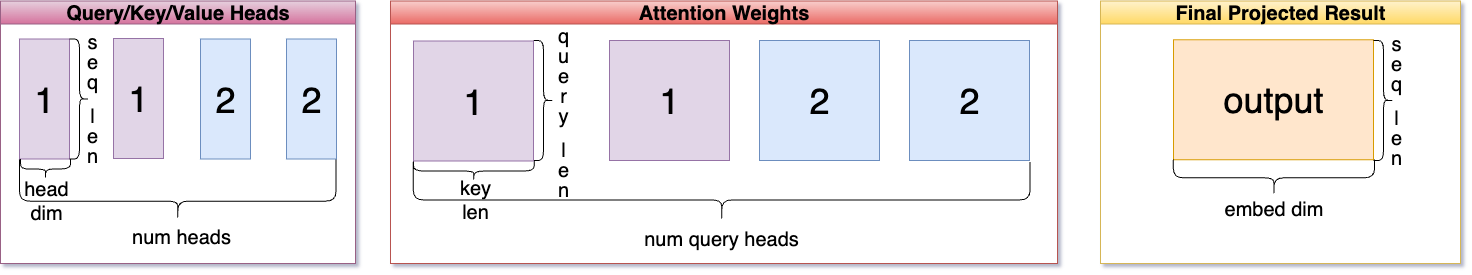}
    \caption{Legend for attention operations diagrams}
    \label{fig:placeholder}
\end{figure}

\begin{figure}
    \centering
    \includegraphics[width=1\linewidth]{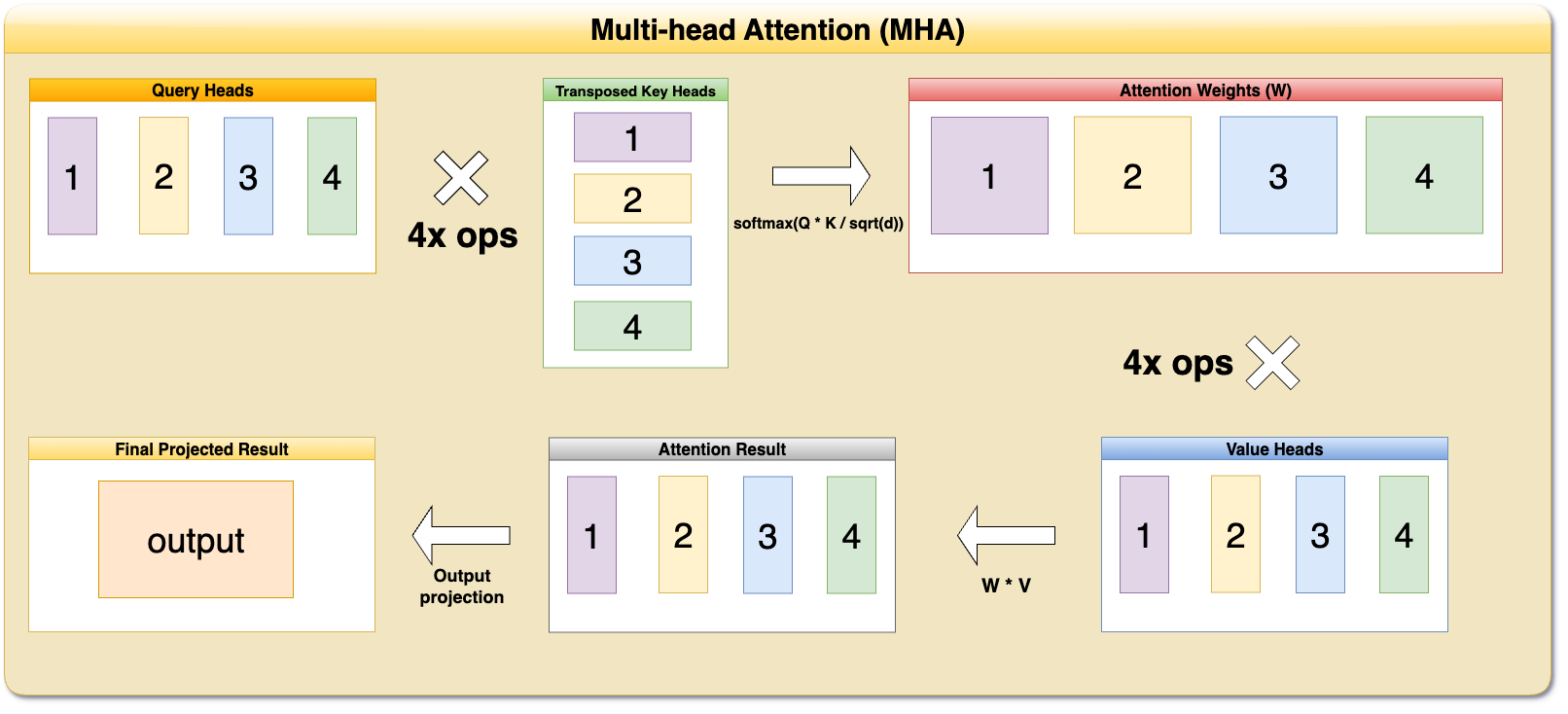}
    \caption{Multi-Head Attention (MHA) operations}
    \label{fig:placeholder}
\end{figure}

\subsection{Multi-Query Attention (MQA): A Radical Solution for Memory Bandwidth}
As Transformer models grew in size and were applied to longer sequences, the cost of autoregressive inference became a critical operational challenge. The MHA design, conceived in an era of smaller models, proved to be inefficient for this specific task. The analysis by Shazeer (2019) identified the memory bandwidth required to load the KV cache as the primary performance limiter on modern accelerators like GPUs and TPUs.

Multi-Query Attention (MQA) was proposed as a direct and aggressive solution to this problem. The architecture is identical to MHA with one crucial difference: a single Key and Value head is shared across all Query heads. This means that while there are still $h$ independent query projections, there is only one key projection and one value projection for the entire layer.

The primary benefit of this design is a dramatic reduction in the size of the KV cache. In MHA, the KV cache for a sequence of length $N$ has a size of $2 \cdot N \cdot h \cdot d_k = 2 \cdot N \cdot d_{\text{model}}$. In MQA, this is reduced to $2 \cdot N \cdot d_k$. The amount of data that needs to be loaded from HBM at each decoding step is therefore reduced by a factor of $h$, the number of heads. This directly translates to a significant speed-up in inference, with reported improvements of up to 12x in certain configurations.

However, this efficiency comes with a trade-off. By forcing all query heads to share the same key and value representations, MQA reduces the model's capacity. This can lead to a degradation in model quality and has been observed to sometimes cause training instability. MQA represents an extreme point on the efficiency-quality spectrum, prioritizing speed above all else.

\begin{figure}
    \centering
    \includegraphics[width=1\linewidth]{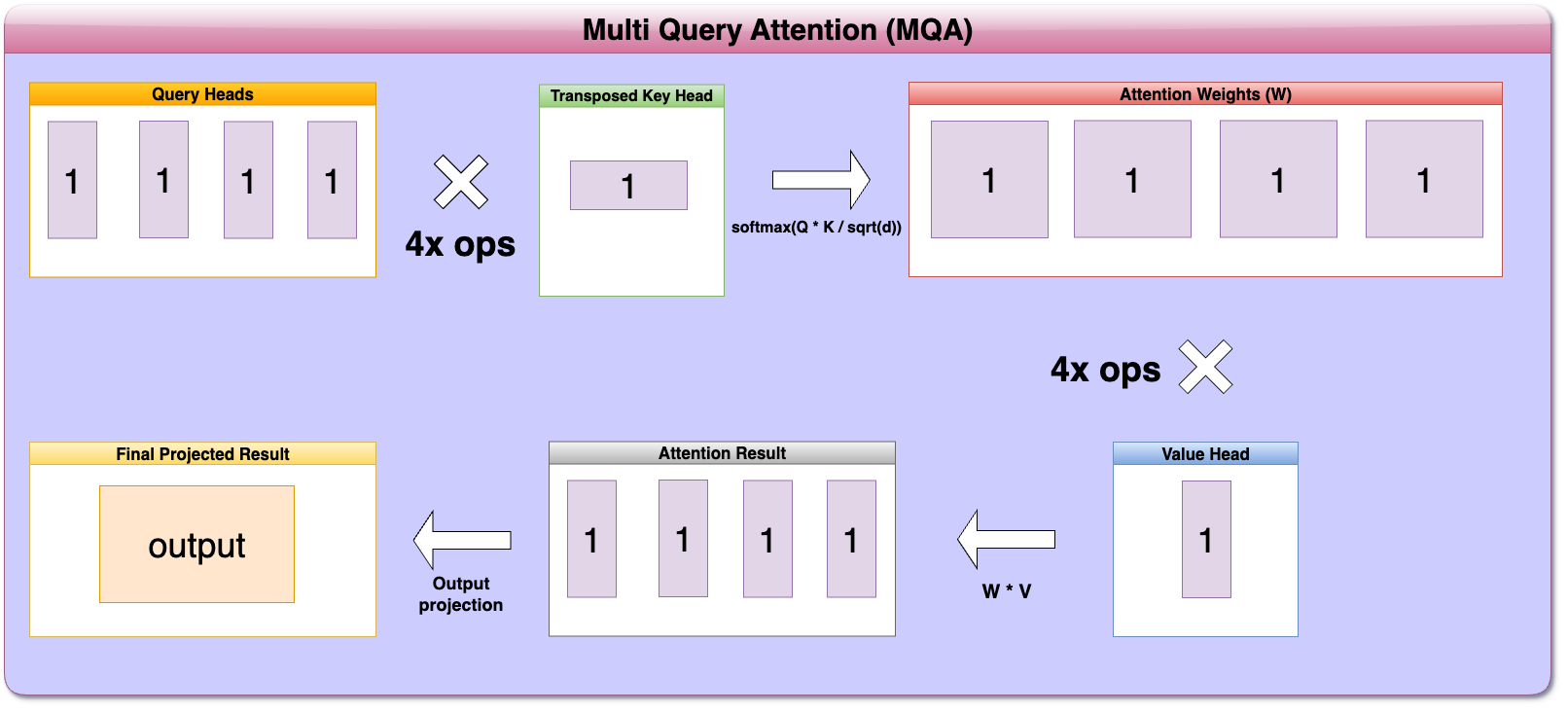}
    \caption{Multi-Query Attention (MQA) operations}
    \label{fig:placeholder}
\end{figure}

\subsection{Grouped-Query Attention (GQA): The Balanced Interpolation}
Grouped-Query Attention (GQA) was introduced by Ainslie et al. (2023) as a way to capture the benefits of MQA while mitigating its drawbacks. GQA is a generalization that elegantly interpolates between the full capacity of MHA and the radical efficiency of MQA.

In GQA, the $h$ query heads are divided into $g$ groups, where $1 \le g \le h$. Each group of $h/g$ query heads shares a single Key and Value head. This architecture provides a tunable parameter, $g$, that allows model designers to control the trade-off between performance and quality.
\begin{itemize}
    \item When $g=h$, each query head has its own key/value head, and GQA becomes equivalent to MHA.
    \item When $g=1$, all query heads share a single key/value head, and GQA becomes equivalent to MQA.
\end{itemize}
By choosing an intermediate value for $g$ (e.g., $g=8$ for a model with $h=32$ query heads), GQA can achieve a substantial reduction in KV cache size and memory bandwidth requirements, leading to inference speeds that are comparable to MQA, while maintaining model quality that is much closer to that of MHA.

\begin{figure}
    \centering
    \includegraphics[width=1\linewidth]{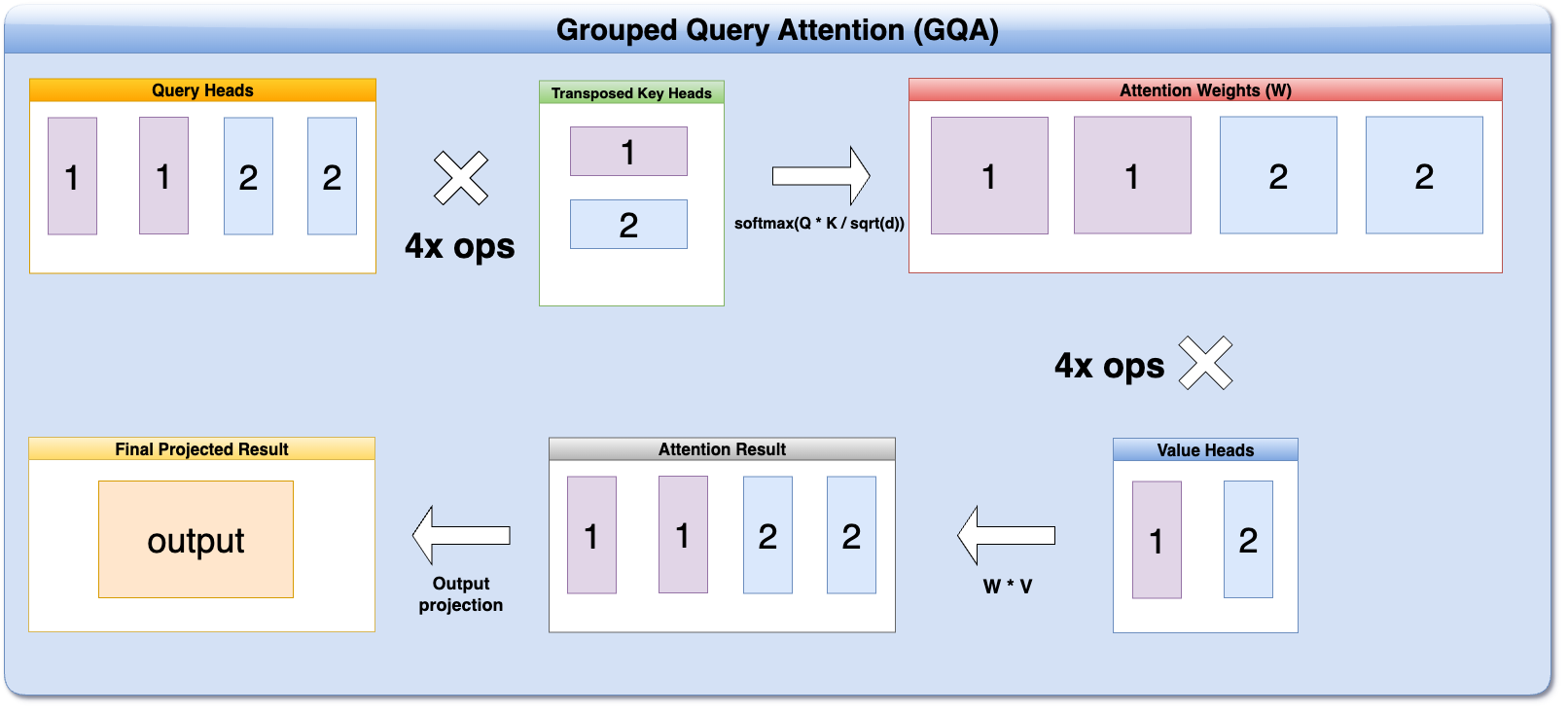}
    \caption{Grouped-Query Attention (GQA) operations}
    \label{fig:placeholder}
\end{figure}

GQA has become a standard in modern LLMs like Qwen2 (Bai et al., 2024), as it achieves most of the inference speed of MQA while maintaining quality much closer to that of MHA.

\subsection{Multi-head Latent Attention (MLA)}
Multi-head Latent Attention (MLA) is a more recent innovation, introduced in the DeepSeek model family (DeepSeek-AI, 2024), that further optimizes the memory bottleneck. While GQA reduces the KV cache by sharing K/V projection matrices, MLA goes a step further by compressing the K and V tensors themselves. It uses learned low-rank projection matrices to map the full K and V tensors into a smaller, latent space before they are stored in the KV cache.

This approach can yield a significantly smaller memory footprint than GQA for the same number of query heads. The trade-off is a slight increase in computation during the projection step. MLA represents the logical endpoint of the memory-optimization trajectory, where the primary goal is to minimize the size of the data transferred from HBM during autoregressive decoding. This reinforces the distinction between memory-centric optimizations and the compute-centric approach of SQA.

The progression from MHA through MQA to GQA and MLA is a clear example of hardware-aware algorithmic design. These architectures are not just abstract mathematical constructs; they are pragmatic engineering solutions tailored to the memory hierarchy of modern GPUs. They directly attack the "memory wall"—the growing gap between compute speed and memory access speed—by minimizing data movement, which is often more energy-intensive and time-consuming than the computation itself. This evolutionary path highlights a critical principle: the most effective deep learning architectures are those that are co-designed with the underlying hardware in mind. SQA continues this trend, but by targeting a different aspect of the hardware: the computational units themselves. While MQA/GQA/MLA optimize for the memory bus, SQA additionally optimizes for the Tensor Cores.

\subsection{Sliding Window Attention (SWA)}
Sliding Window Attention (SWA) is a popular method for approximating full attention to reduce its quadratic complexity (Beltagy et al., 2020). The core idea is to restrict each token's attention computation to a local neighborhood or "window" of a fixed size, k. For a token at position i, it only attends to tokens in the range [i-k/2,i+k/2]. This changes the complexity from $O(N^2 \cdot d_{\text{model}})$ to a much more manageable $O(N \cdot k \cdot d_{\text{model}})$.

The primary drawback of this approach is the introduction of a strong locality bias. The model cannot directly capture dependencies between tokens that are separated by more than the window size. This can be detrimental for tasks requiring long-range reasoning. Architectures like Longformer (Beltagy et al., 2020) address this by combining SWA with a few "global attention" tokens, which are allowed to attend to the entire sequence, creating a hybrid sparse attention pattern. SWA is a complementary mechanism, often used in models that also employ GQA to manage both computational complexity and memory bandwidth.

\section{Sparse Query Attention (SQA)}
Sparse Query Attention (SQA) introduces a new axis of optimization for attention mechanisms. It primarily targets the computational complexity of the attention score calculation, a complementary approach to the memory-centric optimizations of MQA and GQA. It achieves this by reducing the number of query heads. While this is its main innovation, SQA still leverages the reduction of key/value heads to maintain memory efficiency, offering a spectrum of configurations.

Some variants, like Symmetric SQA, may consciously increase the number of K/V heads relative to a GQA baseline to improve quality, making them ideal for full sequence processing and cases where KV cache size is less critical. Other variants, like baseline SQA and Extreme SQA, can be configured to match the memory footprint of GQA, making them suitable for traditional LLMs.

\subsection{The Core Concept: Reversing the GQA Paradigm}
The conceptual foundation of SQA can be understood by asking a simple question: "What if, instead of reducing the number of Key and Value heads as in MQA and GQA, we reduce the number of Query heads?" This line of inquiry leads to a fundamentally different performance profile.

The computational cost of the scaled dot-product attention is primarily driven by the matrix multiplication $QK^T$. The dimensions of this operation are $(N \times d_q) \times (d_k \times N)$, where $d_q$ and $d_k$ are the total dimensions of the query and key projections, respectively. In a standard MHA model, $d_q = d_k = h \cdot d_{\text{head}} = d_{\text{model}}$. The resulting attention score matrix has dimensions $(N \times N)$. This computation is performed for each of the $h$ heads (or in a batched manner), leading to a total complexity proportional to $h \cdot N^2 \cdot d_{\text{head}}$.

MQA and GQA reduce the number of unique key/value projections, but they still require the keys and values to be broadcast or repeated to match the full number of query heads before the attention computation. Therefore, the dimensions of the $Q$ and $K$ matrices entering the dot product remain effectively the same, and the number of FLOPs is not reduced.

SQA takes the opposite approach. By reducing the number of query heads to $H_q < H$, it directly shrinks the dimension of the query matrix $Q$. This results in a smaller number of attention score calculations. As illustrated in the provided architectural diagrams, reducing the number of query heads leads to a proportionally smaller number of attention weight matrices being computed, which in turn reduces the number of value heads that need to be aggregated. This directly translates into a reduction in the total number of floating-point operations.

\subsection{Mathematical Formulation}
Let us formalize the SQA mechanism. We define the following parameters:
\begin{itemize}
    \item $H$: The total number of heads in a comparable MHA baseline model.
    \item $H_q$: The number of query heads in the SQA layer ($1 \le H_q < H$).
    \item $H_{kv}$: The number of key/value heads in the SQA layer ($1 \le H_{kv} \le H_q$).
    \item $d_{\text{model}}$: The hidden dimension of the model.
    \item $d_{\text{head}}$: The dimension of each attention head, typically set to $d_{\text{model}} / H$.
\end{itemize}
Given an input sequence representation $X \in \mathbb{R}^{N \times d_{\text{model}}}$, SQA first projects it into Query, Key, and Value matrices using learned weight matrices $W_Q$, $W_K$, and $W_V$:
\begin{align}
Q &= XW_Q, \quad \text{where } W_Q \in \mathbb{R}^{d_{\text{model}} \times (H_q \cdot d_{\text{head}})} \\
K &= XW_K, \quad \text{where } W_K \in \mathbb{R}^{d_{\text{model}} \times (H_{kv} \cdot d_{\text{head}})} \\
V &= XW_V, \quad \text{where } W_V \in \mathbb{R}^{d_{\text{model}} \times (H_{kv} \cdot d_{\text{head}})}
\end{align}
The resulting matrices $Q$, $K$, and $V$ are then reshaped to separate the head dimension:
\begin{align*}
Q &\in \mathbb{R}^{N \times H_q \times d_{\text{head}}} \\
K &\in \mathbb{R}^{N \times H_{kv} \times d_{\text{head}}} \\
V &\in \mathbb{R}^{N \times H_{kv} \times d_{\text{head}}}
\end{align*}
To perform the attention computation, the number of key and value heads must match the number of query heads. This is achieved by repeating the $K$ and $V$ tensors. Let the repetition factor be $G = H_q / H_{kv}$. The key and value heads are repeated $G$ times along the head dimension to create expanded tensors $K' \in \mathbb{R}^{N \times H_q \times d_{\text{head}}}$ and $V' \in \mathbb{R}^{N \times H_q \times d_{\text{head}}}$. This operation is analogous to the one used in GQA to match key/value heads to query groups.

The scaled dot-product attention is then computed in parallel for each of the $H_q$ query heads using the corresponding (repeated) key and value heads:
\begin{equation}
\text{head}_i = \text{Attention}(Q_i, K'_i, V'_i) \quad \text{for } i = 1, \dots, H_q
\end{equation}
where $Q_i$, $K'_i$, and $V'_i$ are the tensors for the $i$-th head. Finally, the outputs of the $H_q$ heads are concatenated and passed through a final linear projection $W^O \in \mathbb{R}^{(H_q \cdot d_{\text{head}}) \times d_{\text{model}}}$ to produce the final output:
\begin{equation}
\text{SQA}(X) = \text{Concat}(\text{head}_1, \dots, \text{head}_{H_q})W^O
\end{equation}
Note that the output projection matrix $W^O$ maps from a smaller dimension $(H_q \cdot d_{\text{head}})$ back to $d_{\text{model}}$, compared to MHA where it maps from $(H \cdot d_{\text{head}})$.

\subsubsection{Complexity Analysis}
The computational complexity of SQA can be analyzed by focusing on the dominant matrix multiplication steps.
\begin{enumerate}
    \item \textbf{Score Calculation ($QK^T$):} This operation is performed for $H_q$ heads. For each head, the multiplication is between a matrix of shape $(N \times d_{\text{head}})$ and a matrix of shape $(d_{\text{head}} \times N)$. The complexity per head is $O(N^2 \cdot d_{\text{head}})$. Across all $H_q$ heads, the total complexity for score calculation is $O(H_q \cdot N^2 \cdot d_{\text{head}})$.
    \item \textbf{Value Aggregation:} The multiplication of the attention scores (shape $N \times N$) with the value matrix (shape $N \times d_{\text{head}}$) also has a complexity of $O(N^2 \cdot d_{\text{head}})$ per head, for a total of $O(H_q \cdot N^2 \cdot d_{\text{head}})$.
\end{enumerate}
The total complexity of the attention operations in SQA is therefore proportional to $H_q \cdot N^2 \cdot d_{\text{head}}$.

Now, consider the baseline MHA model, where the number of query heads is $H$. Its complexity is proportional to $H \cdot N^2 \cdot d_{\text{head}}$. By comparing the two, we can see that the computational complexity of SQA is a factor of $H_q / H$ relative to MHA. This leads to a theoretical computational speed-up of:
\begin{equation}
\text{Speed-up}_{\text{SQA}} = \frac{\text{Complexity}_{\text{MHA}}}{\text{Complexity}_{\text{SQA}}} = \frac{H \cdot N^2 \cdot d_{\text{head}}}{H_q \cdot N^2 \cdot d_{\text{head}}} = \frac{H}{H_q}
\label{eq:speedup}
\end{equation}
This formal derivation provides the theoretical foundation for SQA's performance benefits. A 50\% reduction in query heads leads to a 2x reduction in computational cost for the attention mechanism.

\subsection{Architectural Variants}
The SQA framework allows for a variety of configurations, enabling a trade-off between computational efficiency and model capacity. Several key variants are proposed and explored in this work:
\begin{itemize}
    \item \textbf{Standard SQA:} This is the most general form, where the number of query heads $H_q$ and key/value heads $H_{kv}$ can be chosen independently, with the constraints that $1 \le H_q < H$ and $1 \le H_{kv} \le H_q$. This flexibility allows for fine-grained control over the model's architecture.
    \item \textbf{Symmetric SQA (sSQA):} This is a specific and compelling configuration where the number of query heads and key/value heads are equal and set to half the total number of heads in the baseline MHA model: $H_q = H_{kv} = H/2$. This variant is designed to achieve a clean 2x computational speed-up over MHA while maintaining a symmetric and balanced capacity for queries and keys/values. It represents a principled reduction in complexity.
    \item \textbf{Extreme SQA (xSQA):} This category includes configurations that push the limits of query head reduction, typically where $H_q \le H/4$. These variants are designed to maximize computational savings and are useful for exploring the lower bounds of required query capacity before model quality degrades significantly. For example, an xSQA variant with $H_q = H/8$ would offer a theoretical 8x speed-up in the attention computation.
\end{itemize}

\begin{figure}
    \centering
    \includegraphics[width=1\linewidth]{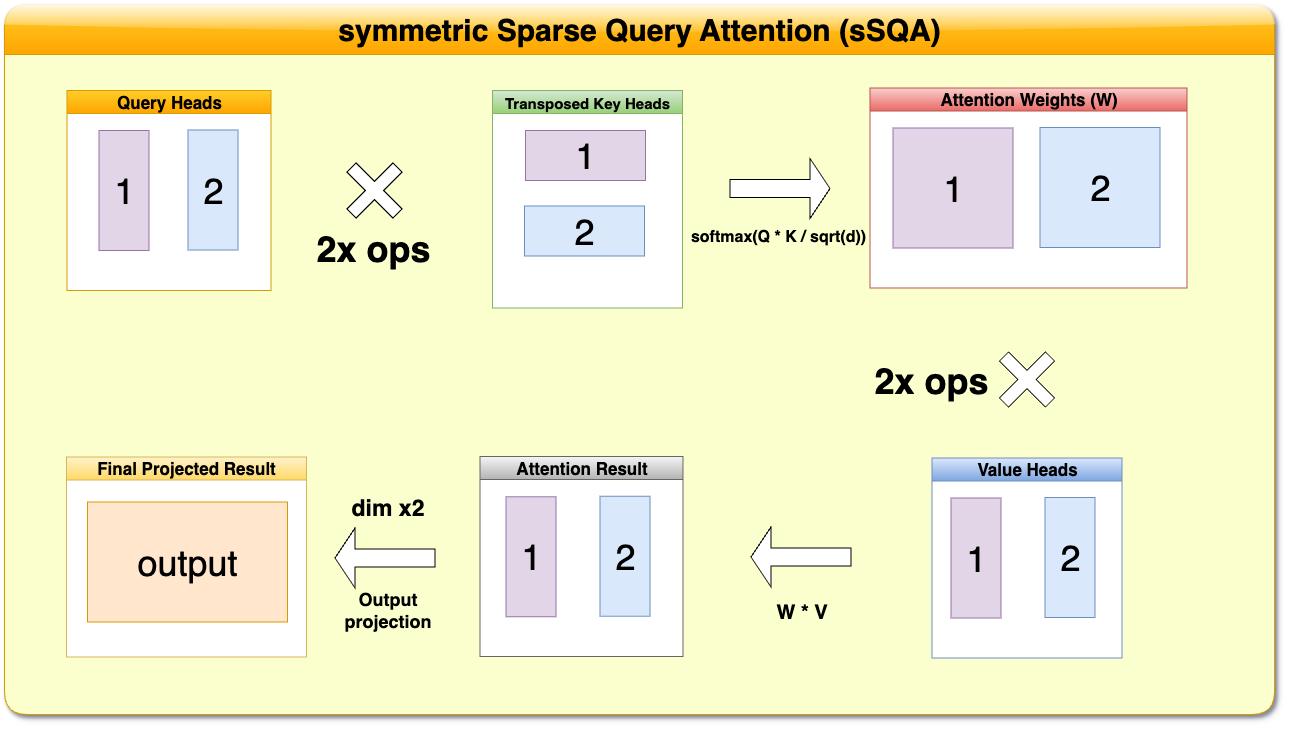}
    \caption{symmetric Sparse Query Attention (sSQA) operations}
    \label{fig:placeholder}
\end{figure}

\begin{figure}
    \centering
    \includegraphics[width=1\linewidth]{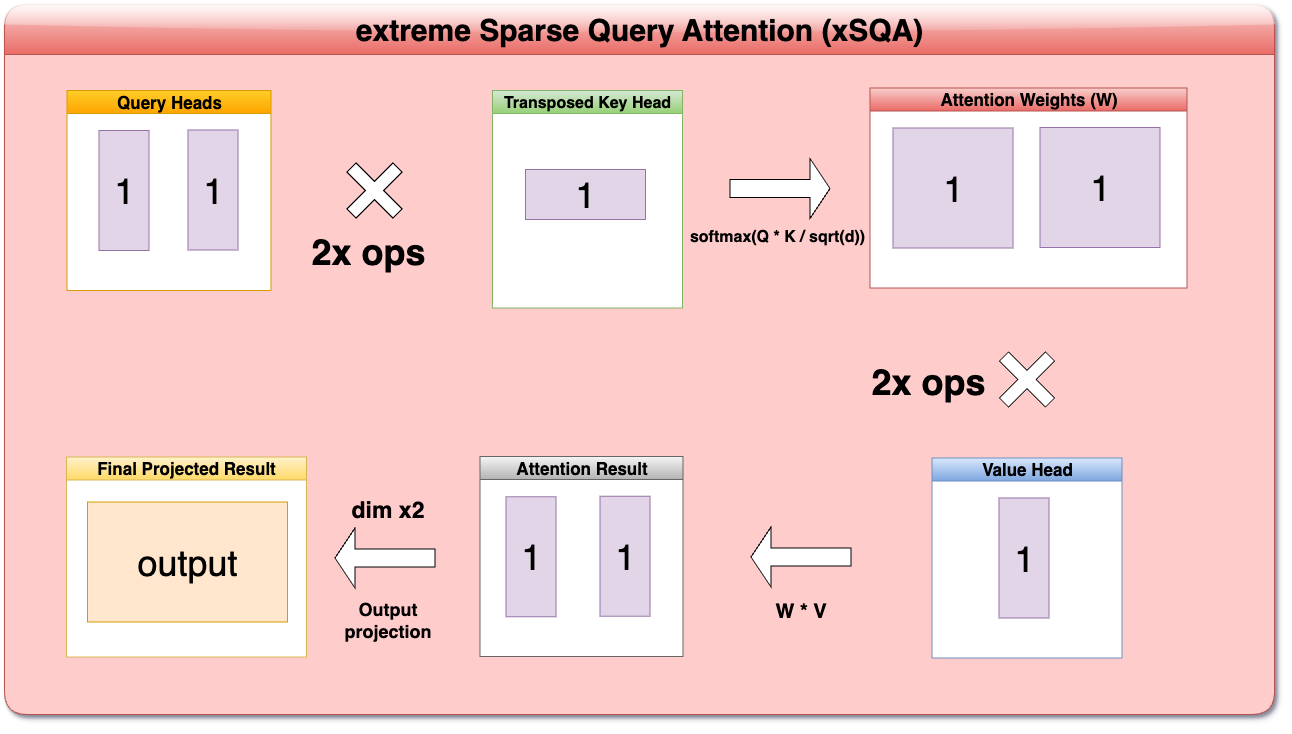}
    \caption{extreme Sparse Query Attention (xSQA) operations}
    \label{fig:placeholder}
\end{figure}

\subsection{Synergy and Composability with Other Mechanisms}
A key advantage of SQA is its architectural simplicity and composability. It can function as a direct, drop-in replacement for any standard attention layer, including MHA, MQA, or GQA, without requiring other changes to the model architecture. This makes it straightforward to integrate into existing models and training pipelines.

Furthermore, SQA is not mutually exclusive with other efficiency mechanisms; it is complementary. Its synergy with Sliding Window Attention (SWA) is particularly noteworthy. A model can be constructed with hybrid "SW-SQA" layers. In such a layer, the attention pattern is first restricted to a local window (the SWA component), and then the attention computation within that window is accelerated by using a reduced number of query heads (the SQA component). This combines the linear complexity scaling of SWA with the constant-factor FLOP reduction of SQA, offering a powerful tool for building highly efficient models for very long sequences. This combination is also allowing to use longer sliding windows with the same efficiency.

\subsection{Comparative Positioning}
To clarify SQA's unique contribution, it is useful to position it relative to the alternatives:
\begin{itemize}
    \item \textbf{vs. GQA/MLA:} SQA reduces FLOPs by shrinking the query matrix. GQA and MLA reduce memory bandwidth by shrinking the KV cache. They target different bottlenecks (computation vs. memory), while some SQA variants have the same influence on memory bottleneck.
    \item \textbf{vs. SWA:} SQA reduces the \textit{cost} of the attention computation. SWA reduces the \textit{scope} of the attention computation (from global to local). SQA is based on structural sparsity, while SWA on spatial. The benefit of the SQA is the access to all tokens, but with partial information about them. They are complementary optimizations that can be used together.
    \item \textbf{vs. SSMs/RetNet:} SQA is an \textit{evolution} of the Transformer's attention block, designed to make it more efficient. SSMs and RetNets are a \textit{replacement} for the attention block, representing a different architectural paradigm.
    \item \textbf{vs. HMT:} SQA is a \textit{layer-level} architectural modification. HMT is a \textit{framework-level} system for managing long contexts via recurrence and memory chunks. They operate at different levels of abstraction.
\end{itemize}

\section{Experiments and Results}

\subsection{Experimental Setup}
Two groups of small-scale models were trained to evaluate SQA against established attention mechanisms.
\begin{itemize}
    \item \textbf{Dense Models:} A set of models with $\sim$10-12M parameters were trained for a single epoch on a 50\% subset of the \texttt{wikimedia/wikipedia} (English) dataset. These models used a hidden dimension of 256, 8 layers, and a baseline of 16 total heads.
    \item \textbf{Mixture-of-Experts (MoE) Models:} A smaller set of MoE models with $\sim$8.5M parameters were trained for 5 epochs on the \texttt{roneneldan/TinyStories} dataset (Eldan, R., \& Li, Y. (2023)). These models used a hidden dimension of 128, 6 layers, and a baseline of 8 total heads.
    \item \textbf{Hardware and Software:} Model quality experiments were conducted on NVIDIA L40S and L4 GPUs. Performance benchmarks were run on a single NVIDIA A100 40GB GPU. The implementation used our internal RxNN framework (v0.1.59), with PyTorch 2.6.0 and Flash Attention 2.7.4.post1 (Dao, T., et al., 2022), and is publicly available in RxNN-Attention (\url{https://github.com/RxAI-dev/rxnn-Attention}) library.
\end{itemize}

\subsection{Model Quality Evaluation}
These experiments provide a preliminary assessment of the impact of reducing query heads on model learning capacity. Due to budget constraints, these evaluations were conducted at a small scale, but they provide valuable evidence of SQA's viability.

\subsubsection{Dense Models ($\sim$10-12M parameters)}
Models were trained with a context size of 1024. The configurations and results are summarized in Table \ref{tab:dense_models}. The SQA variants demonstrate a clear trade-off between performance and quality. Notably, sSQA and SQA achieve validation loss very close to GQA while completing training significantly faster. The xSQA variant is the fastest, with performance still slightly better than MQA.

\begin{table}[h!]
\centering
\caption{Quality and Training Performance of Dense Models}
\label{tab:dense_models}
\begin{tabular}{lcccccc}
\toprule
Model & $H_q$ (of 16) & $H_{kv}$ (of 16) & Val. Loss & Perplexity & Accuracy (\%) & Time (min) \\
\midrule
MHA & 16 & 16 & 1.1976 & 3.3121 & 77.35 & 269 \\
GQA & 16 & 4 & 1.2177 & 3.3794 & 77.12 & 258 \\
MQA & 16 & 1 & 1.2497 & 3.4893 & 76.64 & 261 \\
\textbf{SQA} & \textbf{8} & \textbf{4} & \textbf{1.2272} & \textbf{3.4117} & \textbf{76.97} & \textbf{241} \\
\textbf{sSQA} & \textbf{8} & \textbf{8} & \textbf{1.2201} & \textbf{3.3875} & \textbf{77.05} & \textbf{243} \\
\textbf{xSQA} & \textbf{4} & \textbf{4} & \textbf{1.2428} & \textbf{3.4653} & \textbf{76.74} & \textbf{235} \\
xSMQA & 4 & 1 & 1.2815 & 3.6020 & 76.22 & 235 \\
\bottomrule
\end{tabular}
\end{table}

The models trained in this experiment are available on HuggingFace Hub.
\footnote{
    \url{https://huggingface.co/ReactiveAI/SQAT-m},
    \url{https://huggingface.co/ReactiveAI/sSQAT-m},
    \url{https://huggingface.co/ReactiveAI/xSQAT-m},
    \url{https://huggingface.co/ReactiveAI/xSMQAT-m}
}

\subsubsection{Micro Mixture-of-Experts Models ($\sim$8.5M parameters)}
These smaller MoE models were trained with a short context of 256 tokens. The results in Table \ref{tab:moe_models} show that even with very short sequences, the computational benefits of SQA are noticeable ($\sim$3-4\% faster training time), while the differences in validation loss are minimal. The sSQA configuration is particularly noteworthy, achieving a loss nearly identical to GQA (a $\sim$0.3\% difference) while being 2\% faster.

\begin{table}[h!]
\centering
\caption{Quality and Training Performance of MoE Models}
\label{tab:moe_models}
\begin{tabular}{lcccccc}
\toprule
Model & $H_q$ (of 8) & $H_{kv}$ (of 8) & Val. Loss & Perplexity & Accuracy (\%) & Time (min) \\
\midrule
GQA & 8 & 2 & 1.139 & 3.124 & 70.66 & 398 \\
MQA & 8 & 1 & 1.158 & 3.184  & 70.33 & 399 \\
\textbf{SQA} & \textbf{4} & \textbf{2} & \textbf{1.159} & \textbf{3.187} & \textbf{70.32} & \textbf{387} \\
\textbf{sSQA} & \textbf{4} & \textbf{4} & \textbf{1.142} & \textbf{3.133} & \textbf{70.63} & \textbf{390} \\
\textbf{xSQA} & \textbf{2} & \textbf{2} & \textbf{1.169} & \textbf{3.219} & \textbf{70.12} & \textbf{383} \\
\bottomrule
\end{tabular}
\end{table}

The models trained in this experiment are available on HuggingFace Hub.
\footnote{
    \url{https://huggingface.co/ReactiveAI/GQA-Ref-Micro}, \url{https://huggingface.co/ReactiveAI/MQA-Ref-Micro}, \url{https://huggingface.co/ReactiveAI/SQAT-mm},
    \url{https://huggingface.co/ReactiveAI/sSQAT-mm},
    \url{https://huggingface.co/ReactiveAI/xSQAT-mm}
}

\subsection{Computational Performance Benchmarks}
This section presents the core empirical validation of SQA's primary claim: that it significantly improves performance in compute-bound scenarios. The benchmarks were run on the dense model architecture, measuring the time per step for a forward pass across various sequence lengths.

\begin{table}[h!]
\centering
\caption{Performance Benchmarks for Long Sequence Processing (Time per step in seconds)}
\label{tab:perf_benchmarks}
\begin{tabular}{lccccccc}
\toprule
Seq. Length & xSQA & SQA & sSQA & SWA (128) & MQA & GQA & MHA \\
\midrule
1,024 & \textbf{0.0570} & 0.0642 & 0.0669 & 0.0759 & 0.0760 & 0.0785 & 0.0869 \\
4,096 & \textbf{0.0637} & 0.0750 & 0.0793 & 0.0794 & 0.1001 & 0.1027 & 0.1114 \\
32,768 & \textbf{0.1348} & 0.1991 & 0.2117 & 0.1871 & 0.3612 & 0.3637 & 0.3727 \\
131,072 & \textbf{0.3759} & 0.6308 & 0.6630 & 0.7531 & 1.2530 & 1.2558 & 1.2648 \\
200,000 & \textbf{0.8194} & 1.4116 & 1.4824 & 1.1871 & 2.8555 & 2.8596 & 2.8734 \\
\bottomrule
\end{tabular}
\end{table}

The results provide a clear and compelling confirmation of the theoretical benefits of SQA.
\begin{itemize}
    \item As predicted, the MQA and GQA models show no significant performance improvement over the MHA baseline in this compute-bound setting. Their throughput is virtually identical as sequence length grows.
    \item All SQA variants are significantly faster than MHA, MQA, and GQA, and this performance gap widens dramatically as sequence length increases.
    \item At a sequence length of 200k, the standard SQA model is over 2x faster than GQA ($1.41s$ vs $2.86s$), and the xSQA model is over 3.4x faster ($0.82s$ vs $2.86s$).
    \item For Sliding Window Attention with 128 tokens windows, sliding loop overhead dominates attention calculations.
\end{itemize}
These benchmarks unequivocally demonstrate that SQA is highly effective at reducing computational load and accelerating processing in scenarios dominated by FLOPs. The direct correlation between the reduction in query heads and the increase in throughput validates the core mechanism of SQA.

The results from our small-scale experiments suggest an important scaling dynamic. As demonstrated in Table \ref{tab:perf_benchmarks}, the throughput advantage of SQA over GQA/MHA grows super-linearly with sequence length, as the quadratic computational cost becomes the dominant factor. Conversely, comparing the results across our dense and MoE models (Tables \ref{tab:dense_models} and \ref{tab:moe_models}), the validation loss gap between SQA variants and the GQA baseline remains minimal. This leads to the compelling hypothesis that as models and datasets scale, the representational capacity lost by reducing query heads may be increasingly negligible, while the computational and financial savings become ever more significant. Validating this scaling hypothesis is a critical direction for future work.

\section{Analysis and Discussion}

\subsection{The Performance Profile of SQA: When and Why it Excels}
SQA's advantages are most pronounced in any task that involves parallel, full-sequence processing. Crucially, SQA is not an alternative to sparse patterns like Sliding Window Attention but a complementary technique; SQA can be used to reduce the computational cost of the attention calculated \textit{within} each local window, compounding the efficiency gains and enabling longer sliding windows.
\begin{itemize}
    \item \textbf{Pre-training and Supervised Fine-tuning:} These processes are fundamentally limited by computational throughput. A 2-3x speed-up in the most expensive component of the model, as demonstrated by sSQA and xSQA, translates directly into a substantial reduction in the time and financial cost of training. For organizations training models from scratch, this is a significant advantage.
    \item \textbf{Encoder Architectures:} Any model that relies on an encoder stack, such as for natural language understanding, information retrieval, or as part of a larger system, will benefit directly from SQA. Since encoders process the entire input sequence in parallel, their performance is compute-bound, making SQA an ideal choice.
    \item \textbf{Prompt Processing Phase in LLMs:} For modern LLMs that handle very long contexts, the initial processing of the user's prompt can be a significant source of latency. This "prompt phase" is a parallel, non-autoregressive computation over the entire input sequence. SQA can drastically accelerate this step. For an application with a 100k token context window, speeding up the prompt processing by 2-3x can noticeably improve the user's "time to first token" experience.
\end{itemize}

Conversely, during the \textbf{autoregressive generation phase}, SQA's computational advantage is less impactful. This phase is typically memory-bandwidth-bound, as the model loads the KV cache for each new token. In this regime, the performance of an SQA model will be primarily determined by the size of its KV cache, which is a function of its number of key/value heads ($H_{kv}$). A variant like sSQA with more K/V heads ($H_{kv}=16$) than a comparable GQA model ($H_{kv}=8$) would have a larger KV cache and could be slower during token generation. However, this is a deliberate design choice for quality. SQA configurations can be designed to match the K/V head count of GQA (e.g., an xSQA model with $H_q=8, H_{kv}=8$), thereby matching its memory footprint and performance in memory-bound scenarios. This nuanced behavior does not diminish SQA's value but rather clarifies its optimal application domain.

This leads to a more sophisticated view of model architecture. The traditional, monolithic application of a single attention type throughout a model may be suboptimal. A more principled approach would consider the distinct computational profiles of different phases of operation. For instance, a future LLM architecture could dynamically use an SQA-like mechanism for the compute-bound prompt processing phase and then switch to a GQA-like mechanism for the memory-bound generation phase. This concept of a dynamic "attention profile" represents a promising direction for architectural innovation, moving beyond a one-size-fits-all approach to a more context-aware design.

\subsection{Trade-offs and Broader Implications for LLMs}
When considering the application of SQA to standard, monolithic LLMs like Llama or GPT, a careful analysis of the trade-offs is required. As discussed, deploying an sSQA variant ($H_q=16, H_{kv}=16$) in a model that currently uses GQA with 8 KV heads would double the size of the KV cache. This would likely increase memory consumption during inference and could slow down autoregressive generation.

However, this does not preclude the use of SQA in such models. The variants like xSQA offer a compelling alternative. Consider a xSQA configuration with $H_q=8$ and $H_{kv}=8$. This model would have the \textit{exact same} KV cache size as a standard GQA model with 8 KV heads. Therefore, its performance during memory-bound autoregressive generation would be identical. Yet, it would still benefit from a theoretical 4x computational speed-up ($H/H_q = 32/8 = 4$) during the compute-bound phases of training and prompt processing. This presents a highly attractive configuration: one that matches the state-of-the-art in inference efficiency while offering a substantial acceleration for training and long-prompt ingestion. This demonstrates that SQA is not just a niche solution but a flexible framework that can provide significant value even within the constraints of existing LLM architectures.

\section{Future Work and Extensions}
The promising results from our small-scale experiments strongly motivate the need for validation at a larger scale. The immediate next step for this research will be to apply SQA to a pre-trained, open-source LLM. Specifically, we plan to conduct fine-tuning experiments on a model such as Qwen3-0.6B, where the original GQA layers are replaced with our sSQA and xSQA variants. This will allow for a direct and robust evaluation of SQA's impact on a state-of-the-art architecture and provide clearer insights into the quality-performance trade-off at scale. Beyond this direct validation, several other promising avenues for extending the SQA framework exist:
\begin{itemize}
    \item \textbf{Light SQA (lSQA):} The variants tested in this work focused on aggressive query reduction (50\% or more). It would be valuable to explore "light" SQA configurations with a more modest reduction, for example, setting $H_q = 0.75 \cdot H$. Such a model might offer a 25\% computational speed-up while potentially outperforming GQA on quality metrics, thus finding a new sweet spot on the Pareto frontier.
    \item \textbf{Reverse SQA (rSQA):} An intriguing, though likely less performant, corner of the design space is to have fewer query heads than key/value heads ($H_q < H_{kv}$). In this setup, the query heads would be repeated to match the number of key/value heads. The computational complexity would then scale with $H_{kv}$ instead of $H_q$. While this may not offer a direct performance benefit, exploring its properties could yield deeper insights into the respective roles of queries and keys in the attention mechanism.
    \item \textbf{Flex-SQA:} This direction proposes combining SQA with advanced sparse attention patterns, such as those found in Google's Flex Attention or Longformer. These methods typically combine local (sliding window) attention with a few global attention tokens. Implementing these patterns efficiently, especially with optimized kernels like FlashAttention, can be complex with the asymmetric head configurations of GQA. A symmetric SQA configuration (where $H_q = H_{kv}$) could simplify the implementation and improve the performance of such hybrid patterns. This could enable models with SQA to handle extremely long sequences (e.g., 1M+ tokens) with high efficiency.
    \item \textbf{SW-SQA (Sliding-Window SQA):} A simpler variant of the above would be to apply a standard sliding window attention mechanism on top of an SQA layer. This would combine the FLOPs reduction of SQA with the linear complexity of sliding window attention, potentially creating a highly efficient attention layer for tasks where locality is a strong prior.
\end{itemize}

These potential extensions highlight that SQA is not an endpoint but rather a new building block for designing the next generation of efficient and scalable Transformer models.

\section{Conclusion}
This paper has introduced Sparse Query Attention (SQA), a novel attention mechanism that offers a new and effective strategy for mitigating the computational cost of the Transformer architecture. By challenging the prevailing focus on memory bandwidth optimization and instead targeting the fundamental computational complexity of the attention score calculation, SQA carves out a distinct and valuable niche in the landscape of efficient deep learning.

The core contribution of SQA is its simple yet powerful architectural modification: reducing the number of query heads. This directly reduces the number of floating-point operations required by the attention layer, leading to a theoretical speed-up of $H / H_q$. This work has provided the mathematical formulation for SQA, derived its complexity, and empirically validated its performance benefits. Benchmarks on long sequences conclusively show that SQA can accelerate compute-bound tasks like model training, fine-tuning, and encoding by up to 3x, a domain where existing methods like MQA and GQA provide no advantage. Preliminary experiments suggest that these significant performance gains can be achieved with only a modest and graceful trade-off in model quality.

SQA is particularly well-suited for architectures where computational throughput for full-sequence processing is prioritized over minimizing the autoregressive KV cache. Furthermore, variants like Extreme SQA (xSQA) present a compelling option for standard LLMs, offering the potential for faster training and prompt processing while matching the inference memory footprint of state-of-the-art GQA models.

Ultimately, SQA demonstrates the value of exploring the full design space of attention mechanisms. The optimal architecture is not universal but is instead a function of the specific task, hardware, and performance objectives. By providing a new tool optimized for computational throughput, SQA empowers researchers and practitioners to build more scalable, efficient, and cost-effective models. To facilitate further research, validation, and adoption by the community, the implementation of SQA and its variants has been made publicly available in the RxNN-Attention library (\url{https://github.com/RxAI-dev/rxnn-attention}), in the \texttt{transformers.attention} module. The experiments in this paper were performed in our internal RxNN library (that will be published after Reactive Transformer release), using version 0.1.59 with PyTorch 2.6.0 and Flash Attention 2.7.4.post1.

\begin{multicols}{2}

\end{multicols}

\end{document}